# On the Notion of Cognition

## Carlos Gershenson
### Centrum Leo Apostel, Vrije Universiteit Brussel
### Krijgskundestraat 33, Brussels, 1160, Belgium
### http://student.vub.ac.be/~cgershen
### E- mail: cgershen@vub.ac.be

## Abstract

We discuss philosophical issues concerning the notion of cognition basing ourselves in experimental results in cognitive sciences, especially in computer simulations of cognitive systems. There have been debates on the "proper" approach for studying cognition, but we have realized that all approaches can be *in theory* equivalent. Different approaches model different properties of cognitive systems from different perspectives, so we can only learn from all of them. We also integrate ideas from several perspectives for enhancing the notion of cognition, such that it can contain other definitions of cognition as special cases. This allows us to propose a simple classification of different types of cognition.

## 1. Introduction

When we try to decide wether a system is **cognitive**, we fall into similar problems as the ones we find when deciding wether a system is intelligent, or alive. The notion of cognition has been used in so many different contexts and with so many different goals, that there are several particular definitions, but no general one. We will not try to give a definition of cognition, but we will try to broaden the *notion* of cognition attempting to enhance our understanding of it.

Etymologically, the word 'cognition' comes from the Latin *cognoscere*, which means 'get to know'. We can say that cognition consists in the acquisition of knowledge. Therefore, we can say that *a system is cognitive if it knows something*. Humans are cognitive systems because they *know* how to communicate, build houses, etc. Animals are cognitive systems because they *know* how to survive. Autonomous robots are cognitive systems if they *know* how to perform certain tasks. Does a tree *know* when spring comes because it blossoms? We should better slow down, and first take a look at different approaches for studying cognition.

In classical cognitive science and artificial intelligence (*e.g.* Newell and Simon, 1972; Newell, 1990; Shortliffe, 1976; Fodor, 1976; Pylyshyn, 1984; Lenat and Feigenbaum, 1992), people described cognitive systems as symbol systems (Newell, 1980). However, it seemed to become a consensus in the community that if a system did not used symbols or rules, it would not be cognitive. From this perspective, animals are not cognitive systems because they do not *use* and *have* symbols. Nevertheless, if we open a human brain, we will not find any symbol either. Opposing the symbolic paradigm, the connectionist approach was developed (Rumelhart, *et al.*, 1986; McClelland, *et al.*, 1986), assuming that cognition emerges from the interaction of many simple processing units or neurons. To our knowledge, there has been no claim that "therefore a cognitive system should be able to perform parallel distributed processes, otherwise it is not cognitive". Still, there has been a long discussion on which paradigm is the "proper" one for studying cognition (Smolensky, 1988; Fodor and Pylyshyn, 1988). The behaviour-based paradigm (Brooks, 1986; 1991; Maes, 1994) was developed also opposing the symbolic views, and not entirely different from the connectionist. There have been also other approaches to study cognition (*e.g.* Braitenberg, 1984; Maturana and Varela, 1987; Beer, 2000; Gärdenfors, 2000).

We support the idea that there is no single "proper" theory of cognition, but different theories that study cognition from different <perspectives|contexts> and with different goals. Moreover, we argue that *in theory* any cognitive system can be modelled to an arbitrary degree of precision by most of the accepted theories, but none can do this completely (precisely because they are models). We believe that we will have a less-incomplete understanding of cognition if we use all the theories available rather than trying to explain every aspect of cognition from a single perspective.

In the following section, we compare different approaches used for studying cognition, which leads us to discuss issues about modelling in general in Section 3. In Section 4 we try to broaden the notion of cognition in order to propose one valid in as many contexts as possible. In Section 5 we propose a simple classification of different types of cognition, for then drawing concluding remarks.

## 2. Different Approaches to Cognition: Which One is the Best?

Different approaches used to study cognition can model the same cognitive processes with different complexities, because they make different abstractions, and they have different goals. All of them are simplifying. Symbol systems do not look at subtleties of biological brains; neural networks do not contemplate different types of neurons, particular topologies; neurophysiology can be accused of simplifying molecular interactions, etc. But simplification is a feature of science: we need to ignore details in order to understand phenomena at a particular level. Because of this, we cannot say that any approach is invalid because it simplifies the object of study, since all approaches simplify cognitive processes. Different approaches help us understand different aspects of cognition. Hence, we cannot say that any paradigm is "better" than any other in general, only in particular contexts.

Experimental results ( *e.g.* Gärdenfors, 1994; Seth, 1998; Gershenson, 2002b), indicate that many cognitive processes have been modelled with the same success (to a certain degree) from different paradigms, and also that some paradigms can model the behaviour of other paradigms. We can build a behaviour-based system or a connectionist network which controls agents which can be interpreted to be following rules. But we also can build a rule-based system to mimic the behaviour of "simple" architectures or animals. Seth (1998) has shown how can the performance of evolved Braitenberg (1984) architectures can be interpreted from a behaviourist perspective. We can *describe* the reasoning of a chess player in terms of conditioning, but also the locomotion of bacteria towards benign environments in terms of intentionality (Jonker *et al.*, 2001). Reasoning and behaviour is not something systems *have*, but something we *describe* in them.

For example, a connectionist system cannot be judged "non-cognitive" because "it does not *have* rules". Humans do not *have* rules either. It is a useful way of describing our performance, but we can describe artificial systems in the same way if it is convenient for us.

We can *observe* different artificial systems from different paradigms as performing the same tasks. Gershenson (2002b) showed that we can explain the performance of such a system from a different paradigm. Therefore, we can say that the *cognition of a system is independent of its implementation*. Cognition is rather *observed* than *possessed* by a system, natural or artificial. We *describe* the cognition of a system, but no system has cognition as one of its elements.

Of course, there are several differences among different models from different paradigms. Some models are very easy to implement in software code (also depending on the programmer), others not so much, but occasionally it is a different story if we want to implement an architecture in a real robot. Moreover, if a model works in a simulation and/or robot, it does not mean that animals function in the same way.

Some models are very robust, others would break up quite easily if faced with an unusual situation (but in any case any cognitive system is dependent on its environment). Some models are quite good if we have just practical purposes, if we want things only to work, and this also depends on the experience of the engineer. But if we are interested in using them as explanatory models, the simplicity of their implementation might be secondary, and sometimes even undesired. Also, if we would like to increment or enhance the models, some would need to be redesigned from scratch and others could be easily extended. Also some models would have more ease in adapting to changes of their environment than others, but this does not mean that we cannot adjust different architectures in order to obtain the desired behaviour.

For example, it is not practical to model language development with a Braitenberg architecture or to balance a multi-legged robot using a rule-based system. But these are different aspects of cognition: humans are cognitive systems because they can prove theorems and play chess, but also because they can coordinate their movements in order to communicate and produce effective behaviours[1]. Many people could disagree here, because there are several different uses of the word cognition. We will try to encompass as much as we can, rather than limiting our views and understanding.

## 3. About Models

*"Explanations are for ourselves, not for the explained"*

We can make a more general distinction between models and the modelled with the ontological notion of *relative being* and *absolute being* (Gershenson, 2002a). The absolute being (a-being) is the being which is independent from the observer, and is for and in the whole universe. Therefore, it is unlimited and uncomprehensible, although we can approximate it *as much as we want to*. The relative being (re-being) is the being which is for ourselves, and it is different for each cognizer, and therefore dependent on the observer. It is relative because it depends on the *context* where each cognizer is. This context is different for all cognizers, and even the context of a cognizer can be changing constantly, with his or her notions of what *re-is*. It is limited because cognizers have limits, *i.e.* they cannot *know* an unlimited number of things.

*Everything re-is a generalization of what a-is*. This is because things a-have an unlimited number of properties, but can re-have only a limited number of them, no matter how huge. Therefore, we need to ignore most of these properties (*e.g.* the spins of the electrons while building an aircraft), making a generalization of what things a-are. However, it seems that most of the properties contemplated by different cognizers are the most relevant for their contexts, and there is not much inconvenience in ignoring many properties. But we need to be aware that we will never have a *complete* description of what things a-are, because it would have to be infinite and unlimited.

---

[1]Of course there are different types and degrees of cognition, but we are interested about cognition *in general*.

Different re-beings can generalize an a-being different properties, which might overlap or not. They can also make this generalization observing at different *abstraction levels*. Re-beings can be seen as metaphors of an a-being. Figure 1 shows a diagram of how cognizers can only abstract limited parts of an a-being.

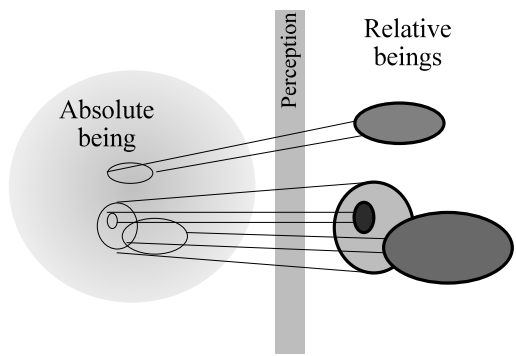

Figure 1. Relative beings as approximations of absolute being.

Returning to models, they can clearly be considered as special cases of re-beings which try to approximate "the real thing" (a-being). All models, by definition, are simplifying. It depends on what we are interested in modelling and how we justify our simplifications that we can judge the suitability of a model[2].

But there is clearly no "best" model outside a context. Some could argue that better models are the ones in which prediction is both maximally accurate and minimally complex. But this is inside the context of information theory.

Moreover, suppose that we are trying to model a function of n dimensions. For any such a function, there is an infinite number of functions of dimension n+1 which determine *exactly* the function we try to model. If we compare two or several of such functions, which one is the "real" one? (Assuming that we cannot know *a priori* how many dimensions the function should have). So, *in theory*, any approach (a different function) could model to a desired level of accuracy a natural cognitive system. In practice different approaches achieve this to different degrees in different perspectives.

We argue that there is no way of determining which re-being is better or worse, good or bad. But one thing we can do,

is to distinguish different degrees of incompleteness. All models are incomplete, but if a model contains several models, it will be less incomplete than those models. This would be valid only in the context of <understanding|explaining>, because in a pragmatic context we would just want a model to work with less effort. But if we try to contain as much <re-beings|models|contexts>, as possible, we will encounter fewer contradictions inside the less incomplete <re-being|model|context>.

Returning even more, to models and architectures of cognition, we can say that not only there is no *general* good model or architecture. And we will have a less incomplete understanding of cognition only if we study it from as many <perspectives|contexts|paradigms> as possible. Each model is abstracting different aspects of what cognition a-is: cognitive behaviour can be described in terms of rules, parallel distributed processing, behaviours, mathematics, etc. And now we can say that all cognitive models are **equivalent** *in the degree that they model the same aspect of cognition*. Some might be more illuminating than others, and so on. Nevertheless, they are just different <ways of | perspectives for> describing the same thing. And this does not mean that cognition is characterized by rules, behaviours, or whatever. *Things do not depend on the models we have of them*.

So, what a-is cognition then?

## 4. About Cognition

Cognition has been studied from a variety of contexts, such as philosophy, artificial intelligence, psychology, dynamical systems theory (Beer, 2000), etc. And it is because of this that in each context cognition is considered with different eyes, and considered to be a different thing. So in different contexts we will be able to define cognition as the manipulation of symbolic representations (Newell, 1990), or as autopoiesis (Stewart, 1996, based on Maturana and Varela, 1980), or as the ability to solve a problem (Heylighen, 1990), or as the ability to adapt to changes in the environment, or as "the art of getting away with it"[3]. We can say that cognition re-is a different thing in different contexts, but can we say what cognition a-is? No, but we can approach to it as much as we want to. The way of achieving this is to make our context as less-incomplete as possible, by containing as many contexts as possible. Therefore, we will not be able to dismiss a model just because it is of a certain paradigm, since all paradigms suffer from limitedness[4]. We can only learn from any model of cognition. We cannot say whether a model is right or wrong *outside a context*. Of course, less-incomplete models will be more robust and will be valid in more contexts. For example, we cannot judge internal representations in a neural context just because these are not <observed|described> at that level.

---

[2]Webb (2001) has discussed several dimensions in which we can make models of animal behaviour more or less close to the modelled: medium, generality, abstraction, level, relevance, structural accuracy, and behaviour match.

[3]This phrase is original of A. Frappé, although referring to intelligence.

[4]*"All ideas are valid in the context they were created".*

We will try to reach a broader notion of cognition basing ourselves on the ideas exposed previously. We can make some general remarks:

- Systems can be **judged** to be cognitive only inside a specific *context*. For example, in a chess-playing context, a bee is not cognitive, but in a navigational context, it is. People agree in contexts, and these are contrasted with experience of a shared world, so we are not in danger of any radical relativism or wild subjectivity.

- Cognition is a **description** we give of systems, not an intrinsic constituent of them, *i.e.* systems do not *have* cognition as an element, *we* observe cognition from a specific context. The cognition of a system does not depend on its implementation.

- If a system performs a **successful** action, *we* can say that it **knows** what to do in that specific situation. This success is tied to a context and to an observer. Therefore, *any system performing a successful action **can** be considered to be a cognitive system*. This is a most general notion of cognition, and other types of cognition and definitions can be applied in different contexts with different purposes without contradicting this notion.

So, a tree *knows* when spring comes because it blossoms, in a specific context (not common in cognitive science, though (yet...)). And a protein *knows* how to become phosphorilized, and a rock *knows* how to fall... if we find a context where this makes sense.

It might seem that we are falling a bit into a language game. *Yes*, but we are victims of the same language game when we speak about human cognition and its limits! *We* are the ones who judge that a tree may know when to blossom, and consider this as knowledge. But this is not different from the process we make when we judge the knowledge of a human or a machine. We can describe human problem solving in terms of behaviour and classical conditioning, but we can also describe biology in terms of epistemology.

We are not insinuating that atoms and humans have the same cognitive abilities, there is a considerable difference in **complexity**, but not in the "essential" nature of cognition (well, the ability to do things "properly" is not entirely essential, since *we* judge this properness). (But for example an oxygen atom *knows* how to bind itself to two hydrogen atoms, and humans do not!).

We can measure this complexity[5], but we should note that this can only be relative to an abstraction level (Gershenson, 2002a). And there are many definitions and measures of complexity, so again there is no "most appropriate" measure outside a context. Moreover, Kolen and Pollack (1994) have shown that complexity is dependent on the observer and how she/he measures a phenomenon.

So, what does cognitive science should study? We would suggest that cognition at all levels, not only at the human, in order to have the broadest notion of cognition. This is not just out of the hat. People already speak about bacterial (Jonker *et al.*, 2001), immunological (Hershberg and Efroni, 2001), plant, animal(Vauclair, 1996; Bekoff, *et al.*, 2002), machine, social, economical cognitions. What is out of the hat is proteic, molecular, atomic, planetary, etc. cognitions. Of course all of this is our interpretation, but if we take "the real thing", what cognition a-is, we humans are not different from any other system. What changes is just how we describe ourselves (and our complexity. This complexity allows us to identify new abstraction levels, and this is very important, but at the end we all are a bunch of molecules, a mass of quarks, and infinitude of nothings...) "How does the immune system *knows* which antigens are foreign of the organism?" is not a question very different from "How do people know when someone is lying?". And research in complex systems (see Bar-Yam (1997) for an introduction) has shown that systems classically considered as cognitive can be modelled with the same models of systems which are classically not considered as cognitive, and also vice versa.

That we are interpreting cognition does not mean that there is no objective notion of cognition. What it means is that it is everywhere, and therefore there is no general way (outside a specific context) to draw a borderline between "cognition" and "non-cognition".

How useful is to describe the behaviour of a particle in terms of cognition, when physics already describes it with a different terminology? It is not about usefulness. We should just realize that cognition, *in essence, a-is* the same for all systems, since it depends on its description. What makes us different is just the complexity degree and the names we use to describe *our* cognition.

## 5. Different Types of Cognition

We can quickly begin to identify different types of cognition, and this will relate the ideas just presented with previous approaches for studying cognition. This does not attempt to be a complete or final categorization, but it should help in understanding our ideas.

We can say that classical cognitive science studies **human cognition**. Of course many disciplines are involved in the study of human cognition, such as neuroscience, psychology, philosophy, classical A. I., etc. Human cognition can be seen as a subset of **animal cognition**, which has been studied by ethologists (*e.g.* McFarland, 1981) and behaviour-based roboticists (*e.g.* Brooks, 1986). But we can also consider the process of life as determined by cognition and vice versa, as the idea of autopoiesis proposes (Maturana and Varela, 1980; 1987; Stewart, 1996), in which we would be speaking about **cognition of living organisms**. Here we would run into the debate of what is considered to be alive, but in any case we can say that biology and artificial life have studied this type of cognition. **Artificial cognition** would be the one exhibited by systems built by us. These can be built as models of the cognition of the living, such as an expert system, an octapod robot, or software agents. But we can also build artificial systems without inspiration from biology which can be considered as cognitive (the thermostat *knows* when it is too hot or too cold). Most of these types of cognition can be considered as **adaptive cognition**, since all living organisms also adapt to modest changes in their environment, but also many artificial and non-living systems. Cybernetics (Wiener, 1948), and more recently certain branches of artificial intelligence and artificial life (*e.g.* Holland, 1992) have studied adaptive systems. We can contain all the previous types of cognition under **systemic cognition**. Complex systems (Bar-Yam, 1997), and general systems theory (Turchin, 1977) can be said to have studied this type of cognition. We cannot think of a more general type of cognition because *something* needs to *exhibit* this cognition, and that something can always be seen as a system. We can see a graphical representation of these types of cognition in Figure 2.

---

[5]A Theaetetic way of measuring cognition could be determined by counting the number of things a system knows...

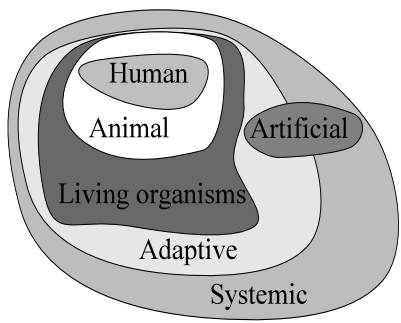

Figure 2. Different types of cognition.

It is curious that cognitions which are considered to be simpler contain the ones considered to be more complex. It seems that this is because when we speak for example about human cognition, we do not see humans as a system, and when we speak for example about cognition in living organisms, we do not think right away about human cognition. We should also note that all types of cognition can be studied at different levels and from different approaches.

This is only one way of categorizing different types of cognition, but there can be several others. One could be by measuring the statistical correlations between the "inputs" and the "outputs" of a cognitive system (if we can identify them). If the outputs can be obtained by pure statistical correlations, then the cognition is simpler than if it requires complex transformation or re-representation of the inputs (Clark and Thornton, 1997; Thornton, 2000). The more transformation the inputs require, the higher and more complex the cognition would have to be. So for example a rock would have low cognition, because if it is on the ground (input), it will stay there (output), and if it is on the air (input), it will fall (output). Now try to do the same predictions/descriptions with a cat, and we can see that they have higher cognition. This categorization is also not universal, but it seems to be useful in several contexts, rather than in a single one.

# 6. Conclusions

In classical cognitive science, it seems that there was the common belief that human cognition *was* a symbol system (Newell, 1990). We believe that the confusion was the following: human cognition *can* be modelled by symbol systems (at a certain level), but this does not mean that human cognition (absolutely) *is* a symbol system. But the same applies to all models. Human cognition (absolutely) is not a parallel distributed processor, nor any other model which we can think about. *Things do not depend on the models we have of them.* And that some aspects of cognition (*e.g.* navigation) are implemented more easily under a certain paradigm, does not mean that natural cognitive systems do it the same way.

Artificial cognitive systems are not cognitive just because they implement a specific architecture. Of course different architectures can be more parsimonious, others more explanatory, others easier to implement, etc.; but this is dependent of the context in which we are modelling, and only in a specific context they can be judged to be cognitive.

Different cognitive models and paradigms can be said to be modelling different aspects of cognition. They are different metaphors, with different goals and from different contexts. Therefore, we will have a less-incomplete view of cognition if we take into account as many paradigms as possible.

There have been several proposed definitions of cognition, in different contexts. We proposed a *notion* which is applicable to all of these contexts and possibly more and encompasses them (although this makes it less practical).

A human doing the same things a simple robot does would be considered cognitive, just because her/his behaviour would be *described* with different terminology. But if the observed processes are the same, we believe that there is no intrinsic cognitive difference related to a specific task between two different systems (*i.e.* functional not material) if they solve the same task in the same context with the same success. This is very similar to Turing's (1950) test for intelligence, only that limited to a context, rather than comparing machines with humans in general. This is why we say that cognition is *observed*. Just as a brain needs a body and an environment (Clark, 1997), a cognitive system also needs an observer.

Someone could say that "real" cognition is given when a system (such as a mature human) is able to explain and understand, and that we are the ones describing other systems, thus gifting them with cognition. We would agree, but even go further: we are the ones describing our own cognition, along with that of any cognitive system. Our cognition does not depend only on our nature, but also on how we <observe|describe> it.

In different contexts, different systems can be judged as cognitive or not, but how we judge them does not change the system, only our description. But contemplating as many systems as cognitive can only enhance our understanding of what cognition is.